\documentclass[letterpaper, 10 pt, conference]{ieeeconf}  % Comment this line out if you need a4paper

\IEEEoverridecommandlockouts                              % This command is only needed if 
\usepackage[T1]{fontenc}
\usepackage{cite}
\usepackage{amssymb,amsfonts}
\usepackage{blindtext}
\makeatletter
\let\NAT@parse\undefined
\makeatother
\usepackage{hyperref}
\usepackage{algorithmic}
\usepackage{graphicx}
\usepackage{textcomp}
\usepackage{mathtools}
\usepackage{changes}
\usepackage[font=footnotesize,labelformat=simple]{subcaption}
\usepackage{algorithm}
\usepackage{xcolor}
\usepackage{color, soul}
\usepackage{amsmath}
\usepackage{booktabs}
\usepackage{multirow}
\usepackage{xstring}
\usepackage{hhline}
\usepackage{kotex}
\usepackage{siunitx}
\usepackage{comment}
\usepackage{physics}
\usepackage{tikz}
\usepackage{cleveref}
\usepackage{gensymb}
\usepackage{threeparttable}
\usepackage{caption}

\definecolor{pr}{RGB}{0, 0, 0}
\definecolor{gl}{RGB}{0, 0, 0}
\definecolor{v3}{RGB}{0,0,0}
\definecolor{v5}{RGB}{0,0,0}
\definecolor{v7}{RGB}{0,0,255}

\definecolor{rv}{RGB}{0,0,0}
\definecolor{rv1}{RGB}{0,0,0}

\title{\LARGE \bf
DreamWaQ: Learning Robust Quadrupedal Locomotion With Implicit Terrain Imagination via Deep Reinforcement Learning
}

\author{I Made Aswin Nahrendra$^{1}$, Byeongho Yu$^{1}$, and Hyun Myung$^{1\ast}$, \textit{Senior Member, IEEE}
\thanks{This work was supported by Korea Evaluation Institute of Industrial Technology (KEIT) grant funded by the Korea Government (MOTIE) (No. \textcolor{rv}{20018216}, ``Development of Mobile Intelligence SW for Autonomous Navigation of Legged Robots in Dynamic and Atypical Environments for Real Application''). The students are supported by BK21 FOUR.}
\thanks{$^{1}$The authors are with the School of Electrical Engineering at Korea Advanced Institute of Science and Technology (KAIST), Daejeon, 34141, Republic of Korea. 
{\tt\footnotesize \{anahrendra, bhyu, hmyung\}@kaist.ac.kr}} \hfill \break
\thanks{$^{\ast}$Corresponding Author: Hyun Myung}\hfill \break
}

\begin{document}

\maketitle
\thispagestyle{empty}
\pagestyle{empty}

%%%%%%%%%%%%%%%%%%%%%%%%%%%%%%%%%%%%%%%%%%%%%%%%%%%%%%%%%%%%%%%%%%%%%%%%%%%%%%%%
\begin{abstract}
Quadrupedal robots resemble the physical ability of legged animals to walk through unstructured terrains. However, designing a controller for quadrupedal robots poses a significant challenge due to their functional complexity and requires adaptation to various terrains. Recently, deep reinforcement learning, inspired by how legged animals learn to walk from their experiences, has been utilized to synthesize natural quadrupedal locomotion. However, state-of-the-art methods strongly depend on a complex and reliable sensing framework. Furthermore, prior works that rely only on proprioception have shown a limited demonstration for overcoming challenging terrains, especially for a long distance. This work proposes a novel quadrupedal locomotion learning framework that allows quadrupedal robots to walk through challenging terrains, even with limited sensing modalities. The proposed framework was validated in real-world outdoor environments with varying conditions within a single run for a long distance.

% resemble the ability of legged animals to walk through unstructured terrains. However, designing a controller for quadrupedal robots poses a significant challenge due to their functional complexity and limited adaptability to various terrains. Meanwhile, legged animals can progressively learn how to walk efficiently and with high adaptability based on their experiences. Recently, deep reinforcement learning (RL), inspired by how legged animals learn to walk, has been utilized to synthesize natural quadrupedal locomotion. However, state-of-the-art methods strongly depend on a complex and reliable sensing framework. Furthermore, prior works that rely only on proprioception have shown a limited demonstration in traversing real-world environments in the long run. This work proposes a novel quadrupedal locomotion learning framework that can traverse challenging terrains, even with limited sensing and in adverse conditions. The proposed framework was validated in real-world outdoor environments with varying conditions within a single run.
\end{abstract}

%%%%%%%%%%%%%%%%%%%%%%%%%%%%%%%%%%%%%%%%%%%%%%%%%%%%%%%%%%%%%%%%%%%%%%%%%%%%%%%%
\section{Introduction}
\noindent In recent years, quadrupedal robots have played an important role in various applications, such as industrial inspection and exploration~\cite{hutter2016anymal,katz2019mini, shin2022design,gehring2021anymal,tranzatto2022cerberus, lee2021qr}. Unlike wheeled mobile robots, quadrupedal robots can traverse unstructured terrains but are relatively difficult to control. Conventional model-based controllers often require a complex pipeline consisting of state estimation, trajectory optimization, gait optimization, and actuator control~\cite{kim2022step, bloesch2013state,hutter2016anymal,gehring2017quadrupedal,bellicoso2017dynamic,jenelten2022tamols, katz2019mini, shin2022design}. Such a complex model-based pipeline
requires considerable human effort for accurate modeling and rigorous parameter tuning. Moreover, the linearized quadrupedal model often limits its performance, hindering its full capability.

Legged animals can efficiently plan their gait by visually perceiving the surrounding terrains. This natural mechanism has inspired many works on training a perceptive locomotion policy via deep reinforcement learning (RL) that can enable a quadrupedal robot to traverse unstructured terrains~\cite{rudin2022learning,miki2022learning,fu2022coupling, yu2021visual}. In these frontier works, the robot is equipped with exteroceptive sensors such as a camera or LiDAR to observe its surroundings. Subsequently, exteroception is used with the controller to plan the robot’s trajectory and gait to traverse through the environment safely.

However, exteroception may not always be dependable. Cameras can malfunction in adverse weather and lighting conditions, and while a 3D LiDAR can be utilized to distinguish ground and traversable regions, accurately estimating the physical characteristics of the terrain remains challenging~\cite{lim2021patchwork,oh2022travel,lee2022patchwork++}. For instance, snow may appear as a solid and passable surface, but it is actually soft and pliable. Additionally, tall grass that appears impassable to a camera can still be easily traversed by legged robots.

% Remove due to irrelevance
%A recent work has shown that terrain properties can be estimated by fusing exteroception and proprioception using a belief encoder~\cite{miki2022learning}. When exteroception is unreliable, the locomotion policy makes decisions based on proprioception. 

% Recently, Miki \textit{et al.}~\cite{miki2022learning} presented a method that can explain the terrain properties by fusing exteroception and proprioception using a belief encoder. When exteroception is unreliable, the locomotion policy makes decisions based on proprioception. 

Meanwhile, proprioceptive sensors, such as an inertial measurement unit (IMU) and joint encoder, are relatively light and robust compared to exteroceptive sensors. Recent works have shown that by combining different proprioception modalities, a quadrupedal robot can learn to estimate its surrounding terrain~\cite{lee2020learning,kumar2021rma, fu2021minimizing,escontrela2022adversarial,margolis2022rapid} and body state~\cite{ji2022concurrent}. However, these works have a limited empirical demonstration for a long-distance operation with various challenging terrains, where legged robots may fail due to high uncertainties and estimation errors.

\begin{figure}[t!]
	\centering 
	\begin{subfigure}[b]{0.48\textwidth}
		\includegraphics[width=1.0\textwidth]{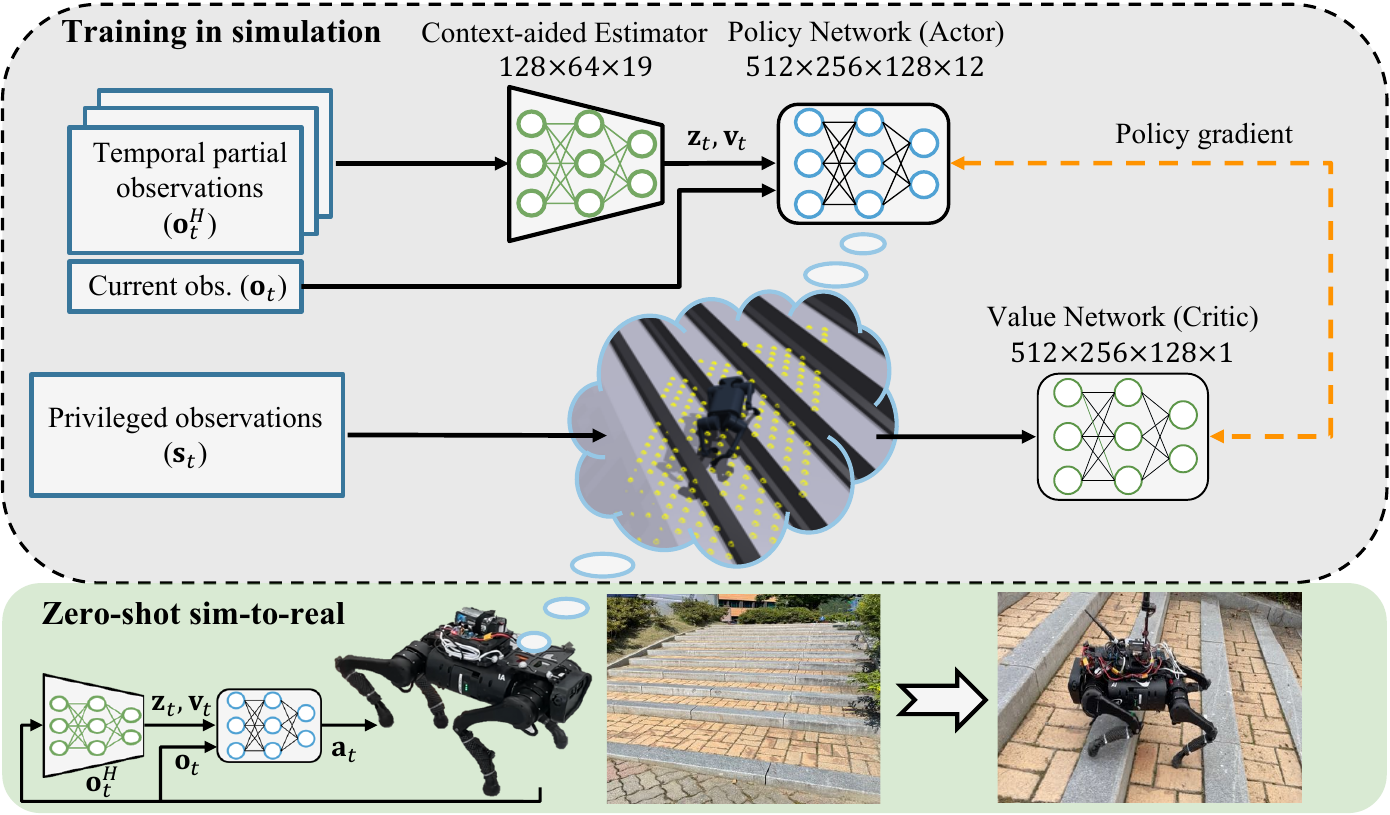}
	\end{subfigure}
	\captionsetup{font=footnotesize}
	\caption{Overview of DreamWaQ. By learning a locomotion policy in a simulation, the robot can walk through challenging terrains such as stairs with zero-shot sim-to-real.}
	\label{figure:overview}
\end{figure}

Estimating the surrounding terrain's properties via proprioception while learning a locomotion policy requires an iterative process~\cite{lee2020learning,kumar2021rma,margolis2022rapid}. The policy needs to understand the terrain properties to learn robust behavior. However, to adequately learn the terrain properties, the robot should be able to walk accordingly and explore a wide spectrum of terrain properties. This dilemma is often called the representation learning bottleneck~\cite{zhang2020learning}, which can hinder optimal policy learning. Therefore, a learning framework that \textcolor{rv}{jointly learn} a robust policy with an accurate environment representation is required.

In this paper, we proposed a framework called \textit{\textbf{Dream} \textbf{Wa}lking for \textbf{Q}uadrupedal Robots} (\textbf{\textit{DreamWaQ}}), that trains a robust locomotion policy for quadrupedal robots with only proprioception via a deep RL algorithm. DreamWaQ trains a locomotion policy to implicitly infer the terrain properties, such as height map, friction, restitution, and obstacles. Consequently, the robot can adapt its gait to walk safely through various terrains. We deployed DreamWaQ on a Unitree A1~\cite{unitreea1} robot to robustly walk through challenging natural and man-made environments.

In summary, the contributions of this work are threefold:
\begin{enumerate}
    \item A novel locomotion learning framework via an asymmetric actor-critic architecture is proposed to implicitly imagine terrain properties using only proprioception.
    \item A context-aided estimator network is proposed to estimate body state and environmental context jointly. Together with the policy, our method outperforms existing learning-based methods.
    \item A robustness and durability evaluation of the learned policy in the real world was conducted through walking in diverse outdoor environments. To the best of our knowledge, this is the first time a Unitree A1, which is significantly smaller that an ANYmal robot, has been demonstrated to sustainably walk on challenging terrain such as hills and yards.\footnote[1]{\label{note1}Project site: \url{https://sites.google.com/view/dreamwaq}}
\end{enumerate}

The remainder of this paper is organized as follows. Section~\ref{section:DreamWaQ} discusses our proposed method thoroughly. Section~\ref{section:experiments} presents the experimental setting, results, and an in-depth comparative analysis of the proposed and baseline methods. Finally, Section~\ref{section:conclusion} concludes this work and briefly discusses directions for future work.

\section{DreamWaQ}\label{section:DreamWaQ}

\subsection{Preliminaries}
In this work, the environment is modeled as an infinite-horizon partially observable Markov decision process (POMDP), defined by the tuple $\mathcal{M}=(\mathcal{S},\mathcal{O},\mathcal{A},d_0,p,r,\gamma)$. The full state, partial observation, and action are continuous, and defined by $\mathcal{\textbf{s}\in\mathcal{S}}$, $\mathcal{\textbf{o}\in\mathcal{O}}$, and $\mathcal{\textbf{a}\in\mathcal{A}}$, respectively. The environment starts with an initial state distribution, $d_0(\textbf{s}_0)$; progresses with a state transition probability $p(\textbf{s}_{t+1}|\textbf{s}_{t},\textbf{a}_{t})$; and each transition is rewarded with a reward function, $r:\mathcal{S}\times\mathcal{A}\to\mathcal{R}$. The discount factor is defined by $\gamma\in[0,1)$. Additionally, in this paper, we define a temporal observation at time $t$ over the past $H$ measurements as $\textbf{o}^H_t=\begin{bmatrix}\textbf{o}_t & \textbf{o}_{t-1} \dots \textbf{o}_{t-H} \end{bmatrix}^T$. We also define a context vector, $\textbf{z}_{t}$, which contains a latent representation of the world state. The context vector is inferred using the method that will be discussed in Section~\ref{section:ce}.

\subsection{Implicit Terrain Imagination}\label{section:implicit}
Recent works have leveraged the teacher-student training paradigm~\cite{chen2020learning}. Although it has been empirically shown that the student policy is as good as the teacher's, behavior cloning (BC) bounds the student policy's performance with the teacher policy~\cite{lee2020learning, kumar2021rma, margolis2022rapid}. Moreover, sequentially training the teacher and student networks is data inefficient~\cite{ji2022concurrent}. The student policy might be unable to explore failure states in which the teacher policy has learned in the early stage of learning using RL. This limitation is because, during BC, the student policy is only provided with good action supervision from the teacher policy. 

\textcolor{rv}{For learning implicit terrain imagination}, we adopted an asymmetric actor-critic architecture~\cite{pinto2017asymmetric}. We discovered that the interplay between the policy and value networks in actor-critic algorithms is sufficient for learning a robust locomotion policy that could implicitly imagine the privileged observations, given partial temporal observations. In DreamWaQ, the policy (actor) receives temporal partial observations, $\textbf{o}^{H}_t$, as the input, while the  value network (critic) receives the full state, $\mathcal{\textbf{s}}_t$, as shown in Fig.~\ref{figure:overview}. In this work, we use $H=5$. Consequently, the data efficiency during training is significantly increased because only one training phase is required. Moreover, the policy can explore all possible trajectories during training, increasing its robustness through generalization. In this work, the policy is optimized using the proximal policy optimization (PPO) algorithm~\cite{schulman2017proximal}.

\subsubsection{Policy Network}
The policy, $\pi_{\phi}(\textbf{a}_t|\textbf{o}_t,\textbf{v}_t,\textbf{z}_t)$ is a neural network parameterized by $\phi$ that infers an action $\textbf{a}_t$, given a proprioceptive observation $\textbf{o}_t$, body velocity $\textbf{v}_t$, and latent state $\textbf{z}_t$. $\textbf{o}_t$ is measured directly from joint encoders and IMU, while $\textbf{v}_t$ and $\textbf{z}_t$ are estimated by a context-aided estimator network (CENet), which will be discussed in Section~\ref{section:ce}. $\textbf{o}_t$ is an $n\times1$ vector defined as follows:
\begin{equation}
  \textbf{o}_t=\begin{bmatrix}\boldsymbol{\omega}_t & \textbf{g}_t & \textbf{c}_t & \boldsymbol{\theta}_t & \boldsymbol{\dot{\theta}}_t & \textbf{a}_{t-1} \end{bmatrix}^T,
  \label{eqn:observation_vector}
\end{equation}
where $\boldsymbol{\omega}_t$, $\textbf{g}_t$, $\textbf{c}_t$, $\boldsymbol{\theta}_t$, $\boldsymbol{\dot{\theta}}_t$, and $\textbf{a}_{t-1}$ are the body angular velocity, gravity vector in the body frame, body velocity command, joint angle, joint angular velocity, and previous action, respectively. 

\subsubsection{Value Network}
The value network is trained to output an estimation of the state value, $V(\textbf{s}_t)$. Unlike the policy, the value network receives the privileged observation, $\textbf{s}_t$,  which is defined as
\begin{equation}
  \textbf{s}_t=\begin{bmatrix}\textbf{o}_{t} & \textbf{v}_{t} & \textbf{d}_{t} & \textbf{h}_{t}\end{bmatrix}^T,
  \label{eqn:privileged_vector}
\end{equation}
where $\textbf{d}_{t}$ is the disturbance force applied randomly on the robot's body and $\textbf{h}_{t}$ is the height map scan of the robot's surroundings as an exteroceptive cue for the value network. In the proposed DreamWaQ, the policy network is trained to implicitly infer $\textbf{d}_{t}$ and $\textbf{h}_{t}$ from proprioception. 

\subsubsection{Action Space}
The action space is a $12\times1$ vector, $\textbf{a}_t$, corresponding to the desired joint angle of the robot. To facilitate learning, we train the policy to infer the desired joint angle around the robot's stand still pose, $\boldsymbol{\theta}_\text{stand}$. Hence, the robot's desired joint angle is defined as 
\begin{equation}
  \boldsymbol{\theta}_\text{des}= \boldsymbol{\theta}_\text{stand} +  \textbf{a}_t.
\end{equation}
The desired joint angles are tracked using a proportional--derivative (PD) controller for each joint.

\subsubsection{Reward Function}
Our reward function closely follows other works~\cite{hwangbo2019learning,lee2020learning,kumar2021rma,rudin2022learning,ji2022concurrent,escontrela2022adversarial} to highlight the effect of DreamWaQ's components instead of reward tuning. The reward function consists of task rewards for tracking the commanded velocity and stability rewards to produce a stable and natural locomotion behavior. The details of the reward function are presented in Table~\ref{table:reward_function}. The total reward of the policy for taking an action at each state is given as:
\begin{equation}
  r_t(\textbf{s}_t,\textbf{a}_t)=\sum r_i w_i,
  \label{eqn:reward}
\end{equation}
where $i$ is the index of each reward, as shown in Table~\ref{table:reward_function}.

The complex reward function for learning a locomotion policy usually includes a motor power minimization term. However, this reward minimizes the overall power without considering each motor's power usage balance. Consequently, in the long run, some motors might overheat faster than others. Therefore, we introduced a power distribution reward to reduce motor overheating in the real world by penalizing motors' power with high variance over all motors used on the robot.

\begin{table}[t!]
% \begin{threeparttable}
\footnotesize
\centering
\captionsetup{font=footnotesize, singlelinecheck=false}
\caption{Reward function elements. $\mathrm{exp(\cdot)}$ and $\mathrm{var(\cdot)}$ are exponential and variance operators, respectively. $(\cdot)^\textbf{des}$ and $(\cdot)^\textbf{cmd}$ indicate the desired and commanded values, respectively. $x, y,$ and $z$ are defined on the robot's body frame, with $x$ and $z$ pointing forward and upward, respectively. $\textbf{g}$, $\textbf{v}_{xy}$, $\omega_\mathrm{yaw}$, $h$, $p_{f,z,k}$, $v_{f,xy,k}$, and $\boldsymbol{\tau}$ are the gravity vector projected into the robot's body frame, linear velocities in the $xy$ plane, yaw rate, body height w.r.t. the ground, foot height, foot lateral velocity, and joint torque, respectively.}
\label{table:reward_function}

\begin{center}
\begin{tabular}{lll}
\hline \hline
Reward           & Equation ($r_i$) & Weight ($w_i$) \\ \hline 
Lin. velocity tracking  & $\exp{-4(\textbf{v}^\mathrm{cmd}_{xy}-\textbf{v}_{xy})^{2}}$ & $1.0$      \\ 
Ang. velocity tracking & $\exp{-4(\omega^\mathrm{cmd}_\mathrm{yaw}-\omega_\mathrm{yaw})^{2}}$ & $0.5$ \\ 
Linear velocity ($z$) & $v^2_z$ & $-2.0$ \\
Angular velocity ($xy$) & $\boldsymbol{\omega}^2_{xy}$ & $-0.05$ \\
Orientation & $|\textbf{g}|^2$ & $-0.2$ \\
Joint accelerations & $\boldsymbol{\ddot{\theta}}^2$ & $-2.5\!\times\!10^{-7}$ \\
Joint power & $\abs{\boldsymbol{\tau}}|\boldsymbol{\dot{\theta}}|$ & $-2\!\times\!10^{-5}$ \\
Body height & $(h^\text{des}-h)^2$ & $-1.0$ \\
Foot clearance & $(p^\text{des}_{f,z,k}-p_{f,z,k})^2\cdot v_{f,xy,k}$& $-0.01$ \\
Action rate & $(\textbf{a}_t - \textbf{a}_{t-1})^2$ & $-0.01$ \\
Smoothness & $(\textbf{a}_t - 2\textbf{a}_{t-1} + \textbf{a}_{t-2})^2$& $-0.01$ \\
Power distribution &$\mathrm{var}(\boldsymbol{\tau}\cdot\boldsymbol{\dot{\theta}})^2$ & $-10^{-5}$ \\ 
\hline \hline
\end{tabular}
\end{center}
% \begin{tablenotes}\footnotesize
% \item[*] 
% \end{tablenotes}
% \end{threeparttable}
\end{table}

% \begin{table}[! htbp]\centering \caption{Summary Statistics}
% \begin{threeparttable}
% \begin{tabular}{l c c c}
% \toprule\midrule
% \thead{Variable} & \thead{Mean}
%  & \thead{Std. Dev.} & \thead{N}\\ \midrule
% a & a & a\tnote{*} & a \\
% \bottomrule\addlinespace[1ex]
% \end{tabular}
% \begin{tablenotes}\footnotesize
% \item[*] Blahblah
% \end{tablenotes}
% \end{threeparttable}
% \label{table2}
% \end{table}

\subsubsection{Curriculum Learning}~\label{section:curriculum}
We utilized a game-inspired curriculum~\cite{rudin2022learning} to ensure progressive locomotion policy learning over difficult terrains. The terrains consisted of smooth, rough, discretized, and stair terrains with ten levels of inclination within $[0\degree,22\degree]$. Furthermore, we found that utilizing the grid-adaptive curriculum~\cite{margolis2022rapid} for low-speed locomotion results in a better and more stable turning that prevents foot tripping.

\subsection{Context-Aided Estimator Network} \label{section:ce}
The policy trained using the method described in Section~\ref{section:implicit} requires $\textbf{v}_t$ and $\textbf{z}_t$ as input, which can be estimated from proprioception. Prior works estimate $\textbf{z}_t$ as the latent variable for understanding terrain properties~\cite{kumar2021rma,fu2021minimizing,margolis2022rapid}. Additionally, estimating $\textbf{v}_t$ using a learned network significantly improves the locomotion policy's robustness~\cite{ji2022concurrent} by eliminating the accumulated estimation drift. 

Motivated by those prior works, we discovered that the interplay between terrain and body state estimates significantly improves body state estimation accuracy. Instead of only explicitly estimating the robot's state, we propose a context-aided estimator network (CENet) architecture to jointly learn to estimate and infer a latent representation of the environment. The advantages of the proposed CENet are: 1) the network architecture is significantly simplified and runs synchronously during inference owing to the shared encoder architecture; 2) the encoder network can jointly learn the robot's forward and backward dynamics via the auto-encoding mechanism, hence, increasing its accuracy.

\begin{figure}[t!]
	\centering 
	\begin{subfigure}[b]{0.48\textwidth}
		\includegraphics[width=1.0\textwidth]{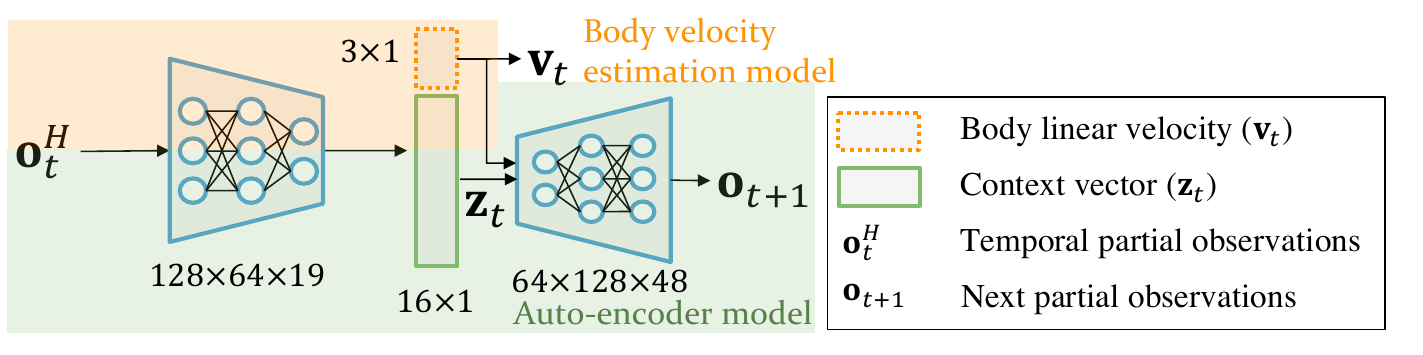}
	\end{subfigure}
	\captionsetup{font=footnotesize}
	\caption{The architecture of CENet consists of a body velocity estimation model and an auto-encoder model that shares a unified encoder. The shared encoder is trained to provide a robust body state and context estimation jointly.}
	\label{figure:CENet}
\end{figure}

CENet consists of a single encoder and a multi-head decoder architecture as shown in Fig.~\ref{figure:CENet}. The encoder network encodes $\textbf{o}^H_t$ into $\textbf{v}_t$ and $\textbf{z}_t$. The first head estimates $\textbf{v}_t$, whereas the second reconstructs $\textbf{o}_{t+1}$. We leveraged a $\beta$-variational auto-encoder ($\beta$-VAE)~\cite{kingma2013auto,higgins2016beta,burgess2018understanding} as the auto-encoder architecture. CENet is optimized using a hybrid loss function, defined as follows:
\begin{equation}
  \mathcal{L}_\text{CE}=\mathcal{L}_{\text{est}}+\mathcal{L}_\text{VAE},
  \label{eqn:ce_loss}
\end{equation}
where $\mathcal{L}_{\text{est}}$ and $\mathcal{L}_\text{VAE}$ are the body velocity estimation and VAE loss, respectively. For explicit state estimation, we employed a mean-squared-error (MSE) loss between the estimated body velocity, $\tilde{\textbf{v}}_t$, and the ground truth, $\textbf{v}_t$, from the simulator as follows:
\begin{equation}
  \mathcal{L}_\text{est}=MSE(\tilde{\textbf{v}}_t,\textbf{v}_t).
  \label{eqn:est_loss}
\end{equation}

The VAE network is trained with the standard $\beta$-VAE loss, which consists of reconstruction and latent losses. We employed MSE for the reconstruction loss and Kullback-Leibler (KL) divergence~\cite{kullback1951information} as the latent loss. The VAE loss is formulated as
\begin{equation}
  \mathcal{L}_\text{VAE} = MSE(\tilde{\textbf{o}}_{t+1},\textbf{o}_{t+1}) + \beta D_\text{KL}(q(\textbf{z}_t|\textbf{o}^H_t)\parallel p(\textbf{z}_t)),
  \label{eqn:vae_loss}
\end{equation}
where $\tilde{\textbf{o}}_{t+1}$ is the reconstructed next observation, $q(\textbf{z}_t|\textbf{o}^H_t)$ is the posterior distribution of the $\textbf{z}_t$, given $\textbf{o}^H_t$. $p(\textbf{z}_t)$ is the context's prior distribution parameterized by a Gaussian distribution.  We chose a standard normal distribution for the prior distribution because all observations are normalized to have a zero mean and unit variance. 

Additionally, bootstrapping from an estimator network during policy network training may increase the sim-to-real robustness of the learned policy~\cite{ji2022concurrent}. However, we discovered that bootstrapping may also harm the policy’s performance because of the large learning noise at the early stage of learning. Therefore, we propose an adaptive bootstrapping (AdaBoot) method that adaptively tunes the bootstrapping probability during training. AdaBoot is controlled by the \textcolor{rv}{coefficient of variation (CV)}, \textcolor{rv1}{i.e., the ratio of the standard deviation to the mean,} of the episodic reward over $m$ domain-randomized environments. The key idea is that bootstrapping is required when the \textcolor{rv}{CV} of $m$ agents’ rewards is small to make the policy more robust against inaccurate estimation. However, it should not bootstrap when the agents have not learned well enough, as indicated by a large \textcolor{rv}{CV} in their rewards. We define the bootstrapping probability for each learning iteration as follows:
\begin{equation}
  p_\text{boot}= 1 - \tanh(CV(\textbf{R})),
  \label{eqn:adaboot}
\end{equation}
where $p_\text{boot}\!\in\![0,1]$ is the bootstrapping probability and $\textbf{R}$ is an $m\!\times\!1$ vector of episodic rewards from $m$ domain-randomized environments. $CV(\cdot)$, and $\tanh(\cdot)$ are coefficient of variation and hyperbolic tangent operations, respectively. $\tanh$ is used to smoothly upper-bounds $CV(\textbf{R})$ to one.

\section{Experiments}~\label{section:experiments}
\vspace{-0.3cm}
\subsection{Compared Methods}~\label{compared_methods}
For a comparative evaluation, we compared the following algorithms with access to proprioceptions only:
\begin{enumerate}
    \item \textbf{Baseline}~\cite{rudin2022learning}: The policy was trained without any adaptation mechanism. 
    \item \textbf{AdaptationNet}~\cite{kumar2021rma,fu2021minimizing}: The policy was trained with an implicit environmental factor encoder using the student-teacher training framework. The policy network consists of 1D convolutional neural network (CNN) layers and multilayer perceptron (MLP) layers.
    \item \textbf{EstimatorNet}~\cite{ji2022concurrent}: The policy was concurrently trained with an estimator network that explicitly estimates the body state without a context estimation.
    \item \textbf{DreamWaQ w/o AdaBoot}: The proposed method without adaptive bootstrapping.
    \item \textbf{DreamWaQ w/ AdaBoot}: The proposed method with adaptive bootstrapping.
\end{enumerate}
All the methods above were trained using the curriculum strategy and reward functions detailed in Section~\ref{section:DreamWaQ}. For a fair comparison, we used the same network architecture and fixed the initial random seeds for all methods. All networks used exponential linear units (ELUs)~\cite{clevert2015fast} as the activation functions for the hidden layers

\subsection{Simulation}
We used the Isaac Gym simulator~\cite{makoviychuk2021isaac} based on the open-source implementation of~\cite{rudin2022learning} to synchronously train the policy, value, and CENet networks for $1,\!000$ iterations. We trained $4,\!096$ agents domain-randomized agents in parallel. The details of the randomized parameters are listed in Table~\ref{table:domain_randomization}. For all algorithms, the policy network was trained using PPO with clipping range, generalized advantage estimation factor, and discount factor of $0.2$, $0.95$, and $0.99$, respectively. The networks were optimized using the Adam optimizer~\cite{kingma2014adam} with a learning rate of $10^{-3}$.

\begin{table}[t!]
\footnotesize
\centering
\captionsetup{font=footnotesize, justification=centering}
\caption{Domain randomization ranges applied in the simulation.}
\label{table:domain_randomization}
\begin{center}
\scriptsize
\begin{tabular}{lccc}
\hline\hline
Parameter                        & Randomization range & Unit \\\hline
Payload                     & $[-1, 2]$ & $\mathrm{~kg}$     \\ 
$K_p$ factor   & $[0.9, 1.1]$ & $\mathrm{~Nm/rad}$    \\ 
$K_d$ factor        & $[0.9, 1.1]$ & $\mathrm{~Nms/rad}$   \\ 
Motor strength factor        & $[0.9, 1.1]$ & $\mathrm{~Nm}$   \\ 
Center of mass shift        & $[-50,50]$ & $\mathrm{~mm}$    \\
\textcolor{rv}{Friction coefficient}        & $[0.2, 1.25]$ & -    \\
\textcolor{rv}{System delay}        & $[0.0, 15.0]$ &  $\mathrm{~ms}$   \\
\hline\hline
\end{tabular}
\end{center}
\vspace{-0.cm}
\end{table}

All training was performed on a desktop PC with an Intel Core i7-8700 CPU @ 3.20 GHz, 32 GB RAM, and an NVIDIA RTX 3060Ti GPU. Training using the DreamWaQ algorithm took approximately one hour to generate data equal to approximately 46 days of training in the real world.

% \begin{table}[t!]
% \centering
% \captionsetup{font=footnotesize, justification=centering}
% \caption{Training Hyperarameters}
% \label{table:simulation_parameters}
% \begin{center}
% \begin{tabular}{lccc}
% \hline\hline
% Parameter                 & Value       \\\hline
% Number of steps per episode                   & $500$   \\ 
% Number of actors & $8,\!192$  \\ 
% Activation function & ELU~\cite{clevert2015fast}\\
% Learning rate & $0.001$\\ 
% Clipping range & $0.2$\\
% Optimizer & Adam~\cite{kingma2014adam}\\
% Discount factor &$0.99$\\
% GAE factor & $0.95$\\

% \hline\hline
% \end{tabular}
% \end{center}
% \vspace{-0.3cm}
% \end{table}

\begin{figure}[t!]
	\centering 
	\begin{subfigure}[b]{0.48\textwidth}
		\includegraphics[width=1.0\textwidth]{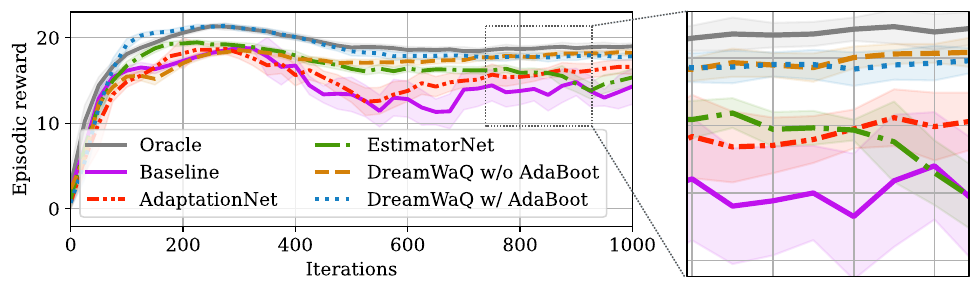}
	\end{subfigure}
	\captionsetup{font=footnotesize}
	\caption{Learning curves of different algorithms. The results shown are obtained from ten different random seeds. \textcolor{rv1}{The curves and shaded regions indicate the mean and standard deviation} of the reward over ten different seeds, respectively. The oracle policy has access to the height map measurement of the robot's surroundings as in~\cite{rudin2022learning}.}
	\label{figure:learning_curves}
\end{figure}

Fig.~\ref{figure:learning_curves} compares the learning curves of DreamWaQ against those of all the other methods for learning the locomotion policy of a Unitree A1 robot. It can be seen that even though EstimatorNet initially has a higher mean episodic reward than AdaptationNet, its performance plummets after more iterations because it encounters more difficult terrains after longer training iterations. Conversely, DreamWaQ consistently outperforms all the other methods. Moreover, despite walking without exteroception, DreamWaQ performs almost as well as the oracle policy that has direct access to the surrounding terrain's height map.

\subsection{Real-World Experimental Setup}
Real-world experiments were conducted using a Unitree A1~\cite{unitreea1} robot. All estimation and control processes were run on an Intel NUC mounted on top of the robot and we used the PyBind interface provided in~\cite{RoboImitationPeng20} to send the joint angle command to the robot. An additional onboard PC with a battery added a payload of approximately 500 g to the robot. During inference, the policy runs synchronously with the CENet at $50~\text{Hz}$. The desired joint angles were tracked using a PD controller with proportional and derivative gains of $K_p=28$ and $K_d=0.7$, respectively at $200~\text{Hz}$.

% \begin{figure}[t!]
% 	\centering 
% 	\begin{subfigure}[b]{0.48\textwidth}
% 		\includegraphics[width=1.0\textwidth]{figures/unitree_a1.jpeg}
% 	\end{subfigure}
% 	\captionsetup{font=footnotesize}
% 	\caption{Unitree A1 robot used in the real-world experiments. (TO BE CHANGED)}
% 	\label{figure:unitree_a1}
% \end{figure}

% \subsection{Latent Space Analysis}
% We analyzed the disentanglement of the latent states encoded by our CENet in comparison with the environmental encoder of RMA. To do it, we collect simulated data with diverse terrains and infer its latent representation using the encoders. Then, we performed dimensionality reduction using the of the latent states using t-SNE~\cite{van2008visualizing} and provide the plot in Fig.~\ref{figure:tsne}

% \begin{figure}[t!]
% 	\centering 
% 	\begin{subfigure}[b]{0.48\textwidth}
% 		\includegraphics[width=1.0\textwidth]{figures/coming_soon.jpeg}
% 	\end{subfigure}
% 	\captionsetup{font=footnotesize}
% 	\caption{t-SNE. (TO BE CHANGED)}
% 	\label{figure:tsne}
% \end{figure}

\subsection{Command Tracking}
\begin{figure}[t!]
	\centering 
	\begin{subfigure}[b]{0.48\textwidth}
		\includegraphics[width=1.0\textwidth]{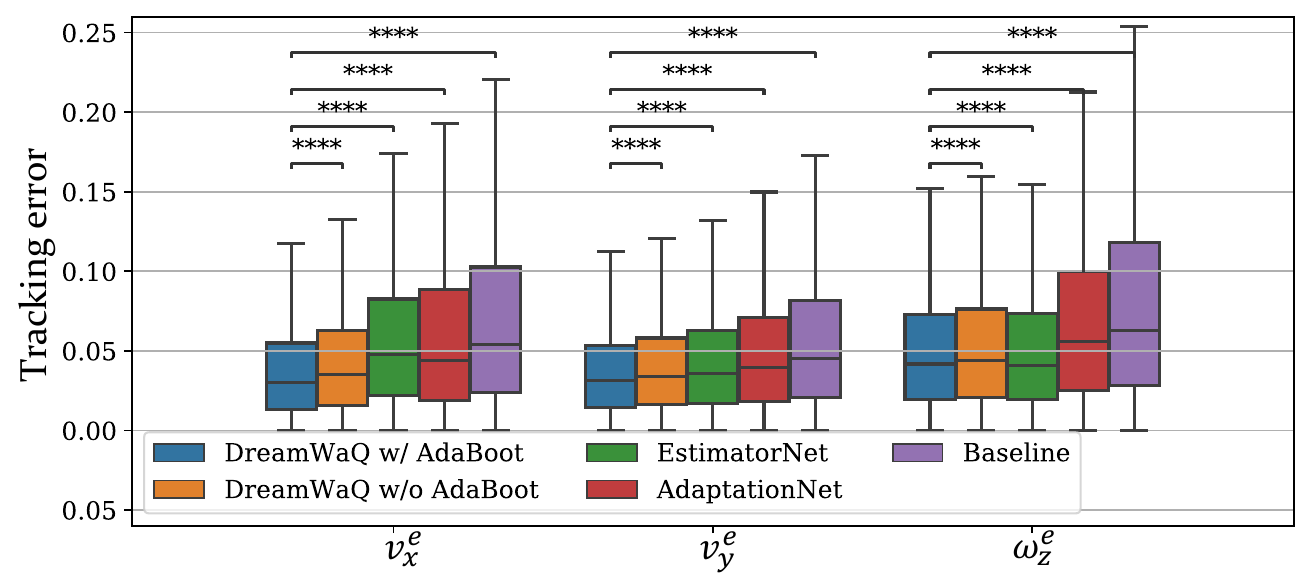}
	\end{subfigure}
	\captionsetup{font=footnotesize}
	\caption{Command tracking error represented as a boxplot. $v^e_{x}$ and $v^e_{y}$ are forward and lateral velocity tracking errors, respectively, measured in $m/s$. $\omega^e_{z}$ is yaw rate tracking error measured in $rad/s$. The $*\!\!*\!\!**$ annotations indicate measurements with $p\text{-value}\!<\!10^{-4}$.}
	\label{figure:tracking_error_boxplot}
\end{figure}

We evaluated the command tracking performance in a \textcolor{rv}{Gazebo simulation} to obtain accurate ground truth. The robot was given random commands for ten minutes, and the commands were \textcolor{rv}{uniformly sampled from $[-1.0,1.0]$} every ten seconds. For fair comparison, random commands were generated using the same random seed for each controller. Each controller was run five times with different random seeds to verify repeatability. We measured absolute tracking error (ATE) as the performance metric and constructed a barplot, as shown in Fig.~\ref{figure:tracking_error_boxplot}. The significance of the improvement obtained by DreamWaQ against other methods was measured using paired $t$-test, as shown in Fig.~\ref{figure:tracking_error_boxplot}, indicating that DreamWaQ consistently outperforms the baselines. Moreover, the proposed AdaBoot method also significantly improved DreamWaQ, owing to its statistical bootstrapping strategy during training.

\subsection{Explicit Estimation Comparison}
\begin{figure}[t!]
	\centering 
	\begin{subfigure}[b]{0.48\textwidth}
		\includegraphics[width=1.0\textwidth]{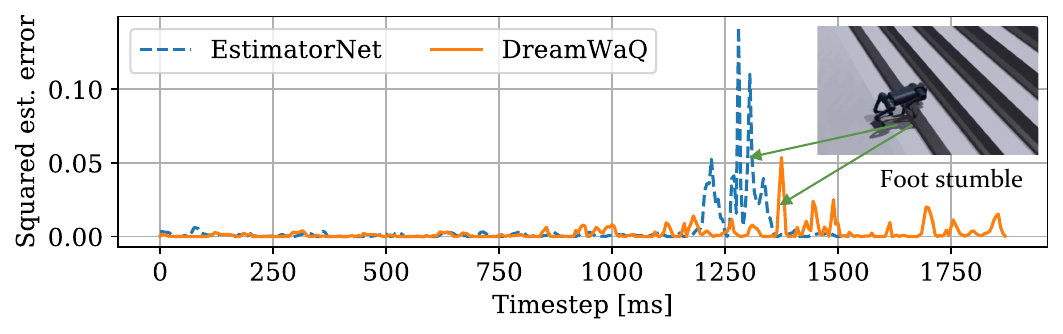}
	\end{subfigure}
	\captionsetup{font=footnotesize}
	\caption{Estimation error of CENet and EstimatorNet. The superiority of CENet is highlighted when the robot's feet stumbled by stairs.}
	\label{figure:stairs_estimation}
\end{figure}

We simulated the robot walking in a stairs environment to compare the CENet with EstimatorNet in terms of their squared estimation error, as shown in Fig.~\ref{figure:stairs_estimation}. In the normal walk condition, CENet shows small errors on the flat terrain, thanks to the forward-backward dynamics learning enabled by the VAE's auto-encoding mechanism. 

The strength of CENet is highlighted when the robot stumbles down the stairs, where the EstimatorNet fails to estimate the body velocity accurately. In severe cases, inaccurate estimation can lead to catastrophic failure. Conversely, the CENet can accurately estimate the body velocity, enabling the robot to climb the stairs safely. We hypothesize that this is made possible by two factors: 1) the forward-backward dynamics learning provides more accurate estimation in all terrains, and 2) using DreamWaQ, the encoder is jointly trained to predict the terrain properties; hence, it can implicitly reason about the terrain properties, which helps in conditioning the explicit estimation.

\subsection{Robustness Analysis}
To test the learned policy's robustness, we perturbed the robot in the simulation with random pushes by applying random velocities with random directions along the $x$, $y$, and $z$ axes of the robot's body frame with a one-second interval until it fell. The random push velocities  were uniformly sampled from $[-\!\textbf{v}^\text{max}_\text{push},\!\textbf{v}^\text{max}_\text{push}]$, where $\textbf{v}^\text{max}_\text{push}\!\geq\!0$ is the maximum push speed. We also measured the survival rate, i.e., the percentage of the robot's survival time within 30 minutes of a random walk. The result of the robustness test is summarized in Table~\ref{table:robustness}.

\begin{table}[t!]
\centering
\captionsetup{font=footnotesize, justification=centering}
\caption{Robustness test. Bold values indicate results with the most robust performance.}
\label{table:robustness}

\begin{center}
\begin{tabular}{lcc}\hline\hline
\multicolumn{1}{l}{Algorithm} & \multicolumn{1}{c}{Max. push (m/s)} & \multicolumn{1}{c}{Survival rate ($\%$)} \\ \midrule
Baseline                           & $0.511 \pm 0.053$                              & $20.51 \pm 6.44$                                   \\ 
AdaptationNet                  & $0.714 \pm 0.096$                                & $82.37 \pm 2.49$                                 \\ 
EstimatorNet                   & $0.871 \pm 0.124$                              & $80.92 \pm 5.73$                                 \\ 
DreamWaQ w/o AdaBoot            & $1.015 \pm 0.121$                             & $90.71 \pm 1.25$                                 \\ 
DreamWaQ w/ AdaBoot             & $\textbf{1.121} \pm \textbf{0.164}$                       & $\textbf{95.23} \pm \textbf{1.61}$                                  \\ 
\hline\hline
\end{tabular}
\end{center}
\end{table}

In all methods, the robot mostly fell when there was a significant change in the command vector, requiring the robot to brake and alter its movement quickly. Nevertheless, DreamWaQ is significantly more robust than all the other methods, as quantitatively verified by the high survival rate and maximum push that it can withstand. The robust performance was achieved through the interplay between accurate estimation and robust policy learning of DreamWaQ. Moreover, the proposed AdaBoot method also increases robustness without sacrificing the base performance. 

\begin{figure*}[t!]
	\centering 
	\begin{subfigure}[b]{0.95\textwidth}
		\includegraphics[width=1.0\textwidth]{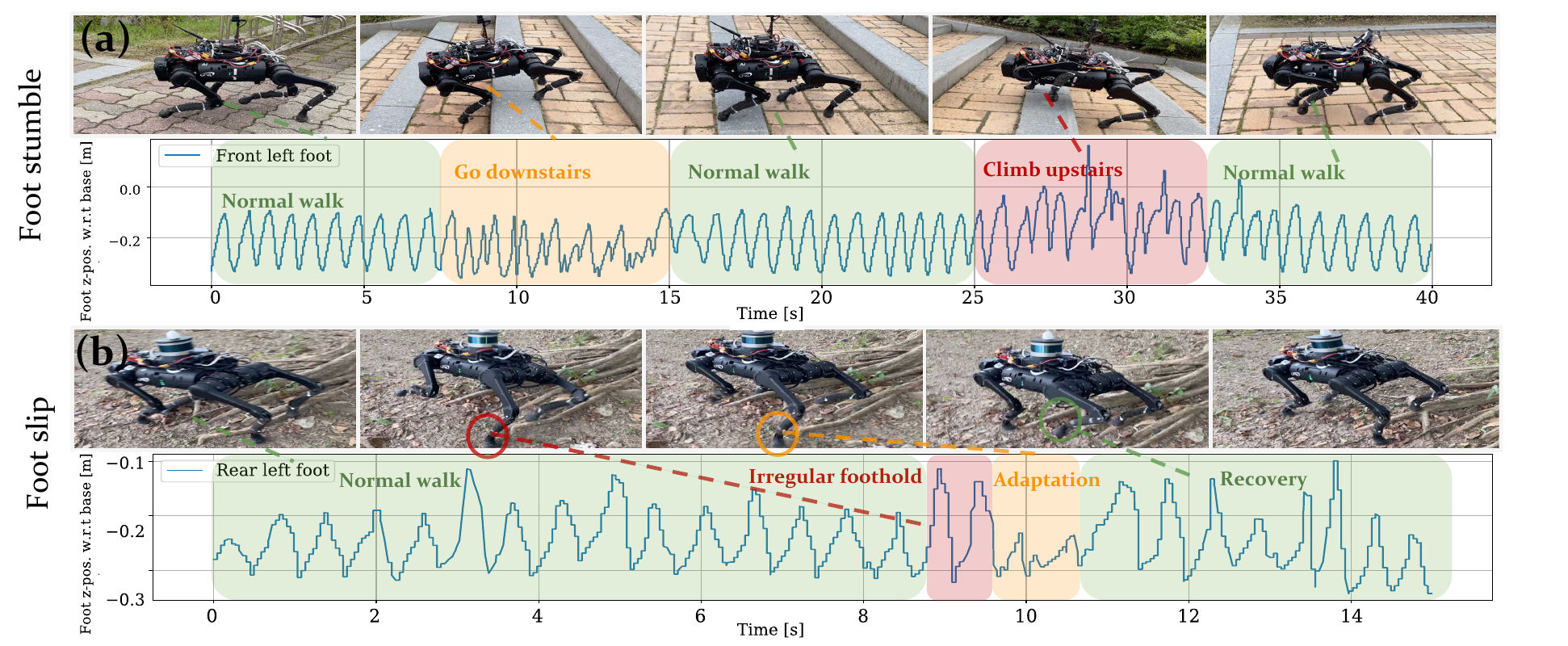}
	\end{subfigure}
	\captionsetup{font=footnotesize}
	\caption{\textcolor{rv}{Foot reflex against uncertainties due to (a) stumbling and (b) slipping in unstructured terrains. Real-time experiment videos are available online\textsuperscript{\ref{note1}}}.}
	\label{figure:foot_reflex}
\end{figure*}

In the real world, DreamWaQ's policy is robust against unstructured terrains. Fig.~\ref{figure:foot_reflex} shows the robot's foot reflex when faced with foot stumbling and slipping. The robot can immediately adapt its gait and stabilize its pose. Owing to the robust and accurate CENet, the robot had no problem in its body velocity estimation and could continue its journey without any performance deterioration.

In Fig.~\ref{figure:foot_reflex}(a), the robot exhibits different gaits for going downstairs and upstairs. When going downstairs, the robot tends to tilt its body closer to the ground and maintain its front foot far from the body, which is a key gait pattern for quickly finding a stable foothold. Meanwhile, the robot adapts its gait for going upstairs by significantly increasing its footsteps. This gait is necessary so that the foot can safely overcome the stairs and find a stable foothold while climbing. Moreover, Fig.~\ref{figure:foot_reflex}(b) shows the adaptation to slipping, where the robot can immediately detect irregular footholds and adapt its gait pattern. Subsequently, the robot tries to recover its normal pattern and continues to walk.

\subsection{Long-Distance Walk}
\begin{figure*}[t!]
	\centering 
	\begin{subfigure}[b]{0.95\textwidth}
		\includegraphics[width=1.0\textwidth]{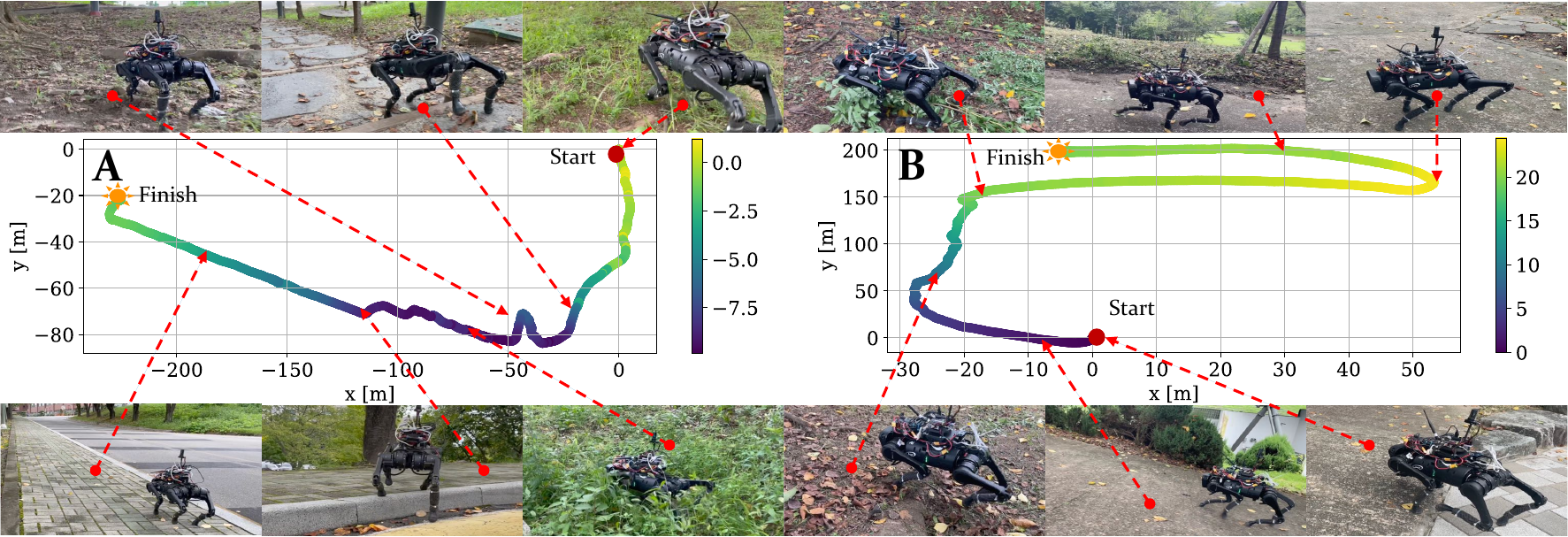}
	\end{subfigure}
	\captionsetup{font=footnotesize}
	\caption{The outdoor trajectory for testing the performance of the DreamWaQ policy was recorded using an RTK--GPS mounted on the robot. Course A consists of many unstructured natural terrains in yards, while course B is a hiking track. \textcolor{rv}{The elevations of both courses relative to the starting point (in $[\mathrm{m}]$) are shown in the color bars.}}
	\label{figure:long_walk}
\end{figure*}

We deployed the robot on two challenging outdoor courses to demonstrate the robustness of DreamWaQ. Course A was an on-campus yard consisting of many slopes and deformable terrains. Course B was an on-campus hill with an elevation gain of up to 22 m. Courses A and B have a total length of 430 m and 465 m, respectively. The details of the courses are shown in Fig.~\ref{figure:long_walk}. The robot's trajectory was measured using a real-time kinematic (RTK) GPS~\cite{rtkgps} with a frequency of 10 Hz, mounted on top of the robot. For complete experiment videos, please refer to the project site\textsuperscript{\ref{note1}}.

\subsubsection{Course A}
The robot was challenged in unstructured natural tracks with various slopes in this course. The robot also encountered thick vegetation that trapped the robot's legs. However, the robot successfully adapted its speed by increasing joint power to overcome the trap.

The most challenging part of this course is walking through stairs and deformable slopes. Thanks to the robustness of the policy and accurate estimation of DreamWaQ, the robot could safely walk through the stairs and slopes. We conducted the experiments not only in dry but also in wet terrain conditions after rainfall. While walking down the stairs, the robot faced slippery stairs. Moreover, the robot's feet stepped deeper to the ground on the slopes because of the mud. Nevertheless, our robot, controlled by DreamWaQ, walked through the wet terrain without any difficulties\textsuperscript{\ref{note1}}.

\subsubsection{Course B}
Course B challenged the robot to climb a moderately high hill. This hiking track consists of man-made asphalt terrain, gravel, and slopes. The experiments were conducted during summer, and the motors heated up quickly. Therefore, we commanded the robot to move slowly to reduce the required torque. Due to the climbing operation, the front legs' motors may easily overheat, and the motor enters the overheat protection mode. Nevertheless, using DreamWaQ, our robot could climb the hill, completing a 465 m trajectory within 10 minutes and reach the hill's summit\textsuperscript{\ref{note1}}.

\section{Conclusion}~\label{section:conclusion}In this work, we introduced DreamWaQ, a robust quadrupedal locomotion framework that enables quadrupedal robots to traverse unstructured terrains by relying solely on proprioception. DreamWaQ showed improved performance compared to existing learning-based controllers, and its robustness was demonstrated on a Unitree A1 robot that walked on hills and unstructured yards for approximately ten minutes. \textcolor{rv}{DreamWaQ's limitation lies in its adaptation mechanism, where it must first hit the obstacles with its legs. Addressing more complex structures, such as high-rise stairs, is a part of our future work, which requires integrating exteroception into the locomotion system for improved gait planning prior to obstacle contact.}

% In this work, we presented DreamWaQ, a robust quadrupedal locomotion framework that enables a quadrupedal robot to walk through traversable obstacles without exteroception. DreamWaQ quantitatively outperforms state-of-the-art learning-based controllers. Subsequently, we demonstrated the robustness of DreamWaQ using a Unitree A1 robot, which walked through a hill and unstructured yards for approximately ten minutes. In the future, we will extend our work to a vision-aided locomotion system to enable autonomous operation in cluttered environments with static and dynamic obstacles.

\bibliographystyle{IEEEtran}
\bibliography{./main,./IEEEabrv}

\begin{thebibliography}{10}
\providecommand{\url}[1]{#1}
\csname url@rmstyle\endcsname
\providecommand{\newblock}{\relax}
\providecommand{\bibinfo}[2]{#2}
\providecommand\BIBentrySTDinterwordspacing{\spaceskip=0pt\relax}
\providecommand\BIBentryALTinterwordstretchfactor{4}
\providecommand\BIBentryALTinterwordspacing{\spaceskip=\fontdimen2\font plus
\BIBentryALTinterwordstretchfactor\fontdimen3\font minus
  \fontdimen4\font\relax}
\providecommand\BIBforeignlanguage[2]{{%
\expandafter\ifx\csname l@#1\endcsname\relax
\typeout{** WARNING: IEEEtran.bst: No hyphenation pattern has been}%
\typeout{** loaded for the language `#1'. Using the pattern for}%
\typeout{** the default language instead.}%
\else
\language=\csname l@#1\endcsname
\fi
#2}}

\bibitem{hutter2016anymal}
M.~Hutter, C.~Gehring, D.~Jud, A.~Lauber, C.~D. Bellicoso, V.~Tsounis,
  J.~Hwangbo, K.~Bodie, P.~Fankhauser, M.~Bloesch, \emph{et~al.}, ``{ANY}mal --
  {A} highly mobile and dynamic quadrupedal robot,'' in \emph{Proc. IEEE/RSJ
  international Conference on Intelligent Robots and Systems (IROS)}, 2016, pp.
  38--44.

\bibitem{katz2019mini}
B.~Katz, J.~Di~Carlo, and S.~Kim, ``Mini cheetah: A platform for pushing the
  limits of dynamic quadruped control,'' in \emph{Proc. IEEE International
  Conference on Robotics and Automation (ICRA)}, 2019, pp. 6295--6301.

\bibitem{shin2022design}
Y.-H. Shin, S.~Hong, S.~Woo, J.~Choe, H.~Son, G.~Kim, J.-H. Kim, K.~Lee,
  J.~Hwangbo, and H.-W. Park, ``Design of {KAIST HOUND}, a quadruped robot
  platform for fast and efficient locomotion with mixed-integer nonlinear
  optimization of a gear train,'' in \emph{Proc. International Conference on
  Robotics and Automation (ICRA)}, 2022, pp. 6614--6620.

\bibitem{gehring2021anymal}
C.~Gehring, P.~Fankhauser, L.~Isler, R.~Diethelm, S.~Bachmann, M.~Potz,
  L.~Gerstenberg, and M.~Hutter, ``{ANY}mal in the field: Solving industrial
  inspection of an offshore {HVDC} platform with a quadrupedal robot,'' in
  \emph{Field and Service Robotics}, G.~Ishigami and K.~Yoshida, Eds.\hskip 1em
  plus 0.5em minus 0.4em\relax Singapore: Springer, 2021, ch.~16, pp. 247--260.

\bibitem{tranzatto2022cerberus}
M.~Tranzatto, T.~Miki, M.~Dharmadhikari, L.~Bernreiter, M.~Kulkarni,
  F.~Mascarich, O.~Andersson, S.~Khattak, M.~Hutter, R.~Siegwart,
  \emph{et~al.}, ``{CERBERUS} in the {DARPA} subterranean challenge,''
  \emph{Science Robotics}, vol.~7, no.~66, p. eabp9742, 2022.

\bibitem{lee2021qr}
E.~M. Lee, D.~Seo, J.~Jeon, and H.~Myung, ``{QR-SCAN}: {T}raversable region
  scan for quadruped robot exploration using lightweight precomputed
  trajectory,'' in \emph{Proc. 21st International Conference on Control,
  Automation and Systems (ICCAS)}, 2021, pp. 957--961.

\bibitem{kim2022step}
Y.~Kim, B.~Yu, E.~M. Lee, J.-H. Kim, H.-W. Park, and H.~Myung, ``{STEP}: State
  estimator for legged robots using a preintegrated foot velocity factor,''
  \emph{IEEE Robotics and Automation Letters}, vol.~7, no.~2, pp. 4456--4463,
  2022.

\bibitem{bloesch2013state}
M.~Bloesch, C.~Gehring, P.~Fankhauser, M.~Hutter, M.~A. Hoepflinger, and
  R.~Siegwart, ``State estimation for legged robots on unstable and slippery
  terrain,'' in \emph{Proc. IEEE/RSJ International Conference on Intelligent
  Robots and Systems (IROS)}, 2013, pp. 6058--6064.

\bibitem{gehring2017quadrupedal}
C.~Gehring, C.~D. Bellicoso, P.~Fankhauser, S.~Coros, and M.~Hutter,
  ``Quadrupedal locomotion using trajectory optimization and hierarchical whole
  body control,'' in \emph{Proc. IEEE International Conference on Robotics and
  Automation (ICRA)}, 2017, pp. 4788--4794.

\bibitem{bellicoso2017dynamic}
C.~D. Bellicoso, F.~Jenelten, P.~Fankhauser, C.~Gehring, J.~Hwangbo, and
  M.~Hutter, ``Dynamic locomotion and whole-body control for quadrupedal
  robots,'' in \emph{Proc. IEEE/RSJ International Conference on Intelligent
  Robots and Systems (IROS)}, 2017, pp. 3359--3365.

\bibitem{jenelten2022tamols}
F.~Jenelten, R.~Grandia, F.~Farshidian, and M.~Hutter, ``{TAMOLS}:
  Terrain-aware motion optimization for legged systems,'' \emph{IEEE
  Transactions on Robotics}, 2022, \url{doi:10.1109/TRO.2022.3186804}.

\bibitem{rudin2022learning}
N.~Rudin, D.~Hoeller, P.~Reist, and M.~Hutter, ``Learning to walk in minutes
  using massively parallel deep reinforcement learning,'' in \emph{Proc.
  Conference on Robot Learning ({CoRL})}, 2022, pp. 91--100.

\bibitem{miki2022learning}
T.~Miki, J.~Lee, J.~Hwangbo, L.~Wellhausen, V.~Koltun, and M.~Hutter,
  ``Learning robust perceptive locomotion for quadrupedal robots in the wild,''
  \emph{Science Robotics}, vol.~7, no.~62, p. eabk2822, 2022.

\bibitem{fu2022coupling}
Z.~Fu, A.~Kumar, A.~Agarwal, H.~Qi, J.~Malik, and D.~Pathak, ``Coupling vision
  and proprioception for navigation of legged robots,'' in \emph{Proc. IEEE/CVF
  Conference on Computer Vision and Pattern Recognition (CVPR)}, 2022, pp.
  17\,273--17\,283.

\bibitem{yu2021visual}
W.~Yu, D.~Jain, A.~Escontrela, A.~Iscen, P.~Xu, E.~Coumans, S.~Ha, J.~Tan, and
  T.~Zhang, ``Visual-locomotion: Learning to walk on complex terrains with
  vision,'' in \emph{Proc. Conference on Robot Learning (CoRL)}, 2021, pp.
  1291--1302.

\bibitem{lim2021patchwork}
H.~Lim, M.~Oh, and H.~Myung, ``Patchwork: concentric zone-based region-wise
  ground segmentation with ground likelihood estimation using a {3D LiDAR}
  sensor,'' \emph{IEEE Robotics and Automation Letters}, vol.~6, no.~4, pp.
  6458--6465, 2021.

\bibitem{oh2022travel}
M.~Oh, E.~Jung, H.~Lim, W.~Song, S.~Hu, E.~M. Lee, J.~Park, J.~Kim, J.~Lee, and
  H.~Myung, ``{TRAVEL}: Traversable ground and above-ground object segmentation
  using graph representation of {3D LiDAR} scans,'' \emph{IEEE Robotics and
  Automation Letters}, vol.~7, no.~3, pp. 7255--7262, 2022.

\bibitem{lee2022patchwork++}
S.~Lee, H.~Lim, and H.~Myung, ``Patchwork++: Fast and robust ground
  segmentation solving partial under-segmentation using {3D} point cloud,''
  \emph{arXiv:2207.11919}, 2022.

\bibitem{lee2020learning}
J.~Lee, J.~Hwangbo, L.~Wellhausen, V.~Koltun, and M.~Hutter, ``Learning
  quadrupedal locomotion over challenging terrain,'' \emph{Science Robotics},
  vol.~5, no.~47, p. eabc5986, 2020.

\bibitem{kumar2021rma}
A.~Kumar, Z.~Fu, D.~Pathak, and J.~Malik, ``{RMA}: Rapid motor adaptation for
  legged robots,'' in \emph{Proc. Robotics: Science and Systems}, 2021.

\bibitem{fu2021minimizing}
Z.~Fu, A.~Kumar, J.~Malik, and D.~Pathak, ``Minimizing energy consumption leads
  to the emergence of gaits in legged robots,'' in \emph{Proc. Conference on
  Robot Learning ({CoRL})}, 2021, pp. 928--937.

\bibitem{escontrela2022adversarial}
A.~Escontrela, X.~B. Peng, W.~Yu, T.~Zhang, A.~Iscen, K.~Goldberg, and
  P.~Abbeel, ``Adversarial motion priors make good substitutes for complex
  reward functions,'' \emph{arXiv:2203.15103}, 2022.

\bibitem{margolis2022rapid}
G.~B. Margolis, G.~Yang, K.~Paigwar, T.~Chen, and P.~Agrawal, ``Rapid
  locomotion via reinforcement learning,'' in \emph{Proc. Robotics: Science and
  Systems}, 2022.

\bibitem{ji2022concurrent}
G.~Ji, J.~Mun, H.~Kim, and J.~Hwangbo, ``Concurrent training of a control
  policy and a state estimator for dynamic and robust legged locomotion,''
  \emph{IEEE Robotics and Automation Letters}, vol.~7, no.~2, pp. 4630--4637,
  2022.

\bibitem{zhang2020learning}
A.~Zhang, R.~McAllister, R.~Calandra, Y.~Gal, and S.~Levine, ``Learning
  invariant representations for reinforcement learning without
  reconstruction,'' in \emph{Proc. International Conference on Learning
  Representations (ICLR)}, 2021.

\bibitem{unitreea1}
\BIBentryALTinterwordspacing
``{U}nitree {A}1,'' accessed on 2022.08.24. [Online]. Available:
  \url{https://m.unitree.com/products/a1}
\BIBentrySTDinterwordspacing

\bibitem{chen2020learning}
D.~Chen, B.~Zhou, V.~Koltun, and P.~Kr{\"a}henb{\"u}hl, ``Learning by
  cheating,'' in \emph{Proc. Conference on Robot Learning (CoRL)}, 2020, pp.
  66--75.

\bibitem{pinto2017asymmetric}
L.~Pinto, M.~Andrychowicz, P.~Welinder, W.~Zaremba, and P.~Abbeel, ``Asymmetric
  actor critic for image-based robot learning,'' in \emph{Proc. Robotics:
  Science and Systems}, 2018.

\bibitem{schulman2017proximal}
J.~Schulman, F.~Wolski, P.~Dhariwal, A.~Radford, and O.~Klimov, ``Proximal
  policy optimization algorithms,'' \emph{arXiv:1707.06347}, 2017.

\bibitem{hwangbo2019learning}
J.~Hwangbo, J.~Lee, A.~Dosovitskiy, D.~Bellicoso, V.~Tsounis, V.~Koltun, and
  M.~Hutter, ``Learning agile and dynamic motor skills for legged robots,''
  \emph{Science Robotics}, vol.~4, no.~26, p. eaau5872, 2019.

\bibitem{kingma2013auto}
D.~P. Kingma and M.~Welling, ``Auto-encoding variational {B}ayes,''
  \emph{arXiv:1312.6114}, 2013.

\bibitem{higgins2016beta}
I.~Higgins, L.~Matthey, A.~Pal, C.~Burgess, X.~Glorot, M.~Botvinick,
  S.~Mohamed, and A.~Lerchner, ``$\beta$ -- {VAE}: Learning basic visual
  concepts with a constrained variational framework,'' in \emph{Proc.
  International Conference on Learning Representations (ICLR)}, 2017.

\bibitem{burgess2018understanding}
C.~P. Burgess, I.~Higgins, A.~Pal, L.~Matthey, N.~Watters, G.~Desjardins, and
  A.~Lerchner, ``Understanding disentangling in $\beta$ -- {VAE},''
  \emph{Advances in Neural Information Processing (NeurIPS) Workshop on
  Learning Disentangled Representations}, 2017.

\bibitem{kullback1951information}
S.~Kullback and R.~A. Leibler, ``On information and sufficiency,'' \emph{The
  Annals of Mathematical Statistics}, vol.~22, no.~1, pp. 79--86, 1951.

\bibitem{clevert2015fast}
D.-A. Clevert, T.~Unterthiner, and S.~Hochreiter, ``Fast and accurate deep
  network learning by exponential linear units ({ELUs}),'' in \emph{Proc.
  International Conference on Learning Representations (ICLR)}, 2016.

\bibitem{makoviychuk2021isaac}
V.~Makoviychuk, L.~Wawrzyniak, Y.~Guo, M.~Lu, K.~Storey, M.~Macklin,
  D.~Hoeller, N.~Rudin, A.~Allshire, A.~Handa, \emph{et~al.}, ``Isaac {G}ym:
  High performance {GPU}-based physics simulation for robot learning,''
  \emph{Advances in Neural Information Processing Systems, Track on Datasets
  and Benchmarks}, 2021.

\bibitem{kingma2014adam}
D.~P. Kingma and J.~Ba, ``Adam: A method for stochastic optimization,'' in
  \emph{Proc. International Conference on Learning Representations (ICLR)},
  2015.

\bibitem{RoboImitationPeng20}
X.~B. Peng, E.~Coumans, T.~Zhang, T.-W.~E. Lee, J.~Tan, and S.~Levine,
  ``Learning agile robotic locomotion skills by imitating animals,'' in
  \emph{Robotics: Science and Systems}, 07 2020.

\bibitem{rtkgps}
\BIBentryALTinterwordspacing
``{H-RTK} {F9P} {H}elical {GPS},'' accessed on 2022.09.02. [Online]. Available:
  \url{http://www.holybro.com/product/h-rtk-f9p/}
\BIBentrySTDinterwordspacing

\end{thebibliography}

\end{document}